\documentclass[conference]{IEEEtran}
\IEEEoverridecommandlockouts 
\usepackage{amsmath,amssymb,amsfonts}
\usepackage{bm}               
\usepackage{graphicx}
\usepackage{booktabs}    
\usepackage[table]{xcolor}   
\definecolor{oursrow}{HTML}{EAF2FB}
\definecolor{privrow}{HTML}{F2F2F2}
\usepackage{multirow}
\usepackage{subcaption}\usepackage{booktabs}  
\usepackage{multirow}  
\usepackage[table]{xcolor}  
\usepackage{algorithm}
\usepackage{algpseudocode}

\usepackage{xcolor}
\usepackage{hyperref}
\hypersetup{colorlinks=true,linkcolor=blue,citecolor=blue,urlcolor=blue}
\usepackage{cite}       
\newcommand*\circled[1]{\textcircled{\raisebox{-0.9pt}{#1}}}

\title{Causeway: Restoring Task Accessibility for Instruction Switching in VLA Policies }

\author{Qingzi Wang$^{1}$, Kaixi Feng$^{1}$, Guangyao Shi$^{2}$, Xiyang Wu$^{1}$, Ang Li$^{1}$, Dinesh Manocha$^{1}$%
\thanks{$^{1}$University of Maryland, College Park, MD, USA.}%
\thanks{$^{2}$University of Southern California, Los Angeles, CA, USA.}%
}

\begin{document}
\maketitle

\begin{abstract}
Vision-language-action (VLA) policies can execute many tasks from standard initial states, yet a new instruction may fail after another task has altered the robot’s physical state. 
We study instruction
switching, where a new task is issued during or after the
execution of a different one. We observe that a target task that is
reliably completed from its standard initial states can become
inaccessible from states produced by a preceding task. We call
such states \emph{task islands}. 
We propose \emph{Causeway}, a training-free inference-time
intervention. Given the current state and a re-entry pose for the
target task, Causeway back-propagates through the frozen decoding
computation and applies a state-directed write within the
action-stream representation. The VLA  decodes the return
motion itself, without parameter updates, a new action head, or
external
action generation. Across 71 cross-object pairs, three
switch timings, and three VLA architectures on LIBERO-Goal~\cite{liu2023libero}, Causeway raises bare-switch success from 3--26\% to 47--65\%
and increases the rate of reaching the handoff neighborhood by
42--72 percentage points across models. Additional experiments
on LIBERO-Object and a real xArm platform show that the recovery extends
beyond the main LIBERO-Goal setting, both in simulation and on a robot.

\end{abstract}

\section{Introduction}

Vision–Language–Action (VLA) policies leverage pre-trained multimodal representations to execute diverse manipulation tasks specified via natural language~\cite{brohan2023rt2,kim2024openvla,black2024pi0,nvidia2025groot}. However, standard benchmarks evaluate tasks from fixed initial distributions~\cite{liu2023libero}. In sequential deployment, completing one task alters the robot and scene state, often rendering subsequent tasks inaccessible. Because resetting between instructions discards progress, practical competence requires that tasks remain accessible
from the states that preceding tasks leave behind.

We study \emph{instruction switching}, in which a policy begins task $A$ and permanently receives instruction $\ell_B$ at an intermediate or terminal state. Although the policy may complete $B$ reliably from its nominal initial states, it can fail to initiate $B$ from states produced by $A$. We call these states \emph{task islands}. Their occurrence depends on the target task and physical state, not simply on elapsed execution time.

Task islands are closely related to the state-dependent steerability studied by ReSteer~\cite{chen2026resteerquantifyingrefiningsteerability}. ReSteer measures post-switch success across intermediate states, attributes limited steerability coverage to insufficient task contrast in the training data, and improves the policy through stage-matched transition generation and self-refining behavioral cloning. We address a complementary problem: online recovery after a bare instruction switch has failed. Rather than generating transition data or updating the policy, we study and recover from inaccessible states in frozen VLAs.

\noindent {\bf Main Results:}
We find that task accessibility can decline near contact and grasp events (Sec.~\ref{sec:escape}). At these states, a new instruction continues to alter the text representation, but its effect on the decoded action weakens substantially (Sec.~\ref{sec:gating}). The action-stream representation nevertheless remains locally steerable by a state-directed gradient.

We introduce Causeway, a training-free inference-time intervention. 
Our formulation addresses the failure by guiding the robot toward a demonstration-supported re-entry region for the target task. Given the current state and a re-entry pose derived from target-task demonstrations, we define a target-directed action objective, differentiate it through the frozen decoder, and add the resulting direction to the action-stream representation. Our frozen policy generates the recovery actions, while proprioceptive thresholds govern intervention and handoff. Causeway neither updates model parameters nor replaces policy control with a new action head or external action generation. Moreover, it does not teach the VLA a new task; rather  it restores access to an existing capability. Our contributions are threefold:
\begin{enumerate}
\item We characterize task islands through target-relative state compatibility and connect post-switch failure to demonstration-supported geometry (Sec.~\ref{sec:state}), behavioral escape maps (Sec.~\ref{sec:escape}), and attenuation of instruction influence (Sec.~\ref{sec:gating}).
\item We introduce Causeway, a training-free representation intervention with component-wise pose steering (Sec.~\ref{sec:write}) and state-dependent engagement and handoff (Sec.~\ref{sec:flow}). Our method applies to parallel, flow-matching, and diffusion-based VLA decoders.
\item 
Across 71 cross-object task pairs, three switch timings, and three VLA architectures on LIBERO-Goal ($\pi_{0.5}$, OpenVLA-OFT, and
GR00T~N1.5), Causeway raises success from 3--26\% to 47--65\% (Sec.~\ref{sec:exp}). Experiments on LIBERO-Object and a real xArm platform (Sec.~\ref{sec:real}) further test its generality.
\end{enumerate}



\nocite{li2025switchvlaexecutionawaretaskswitching}

\begin{figure*}[t]
\centering
\includegraphics[width=0.8\textwidth]{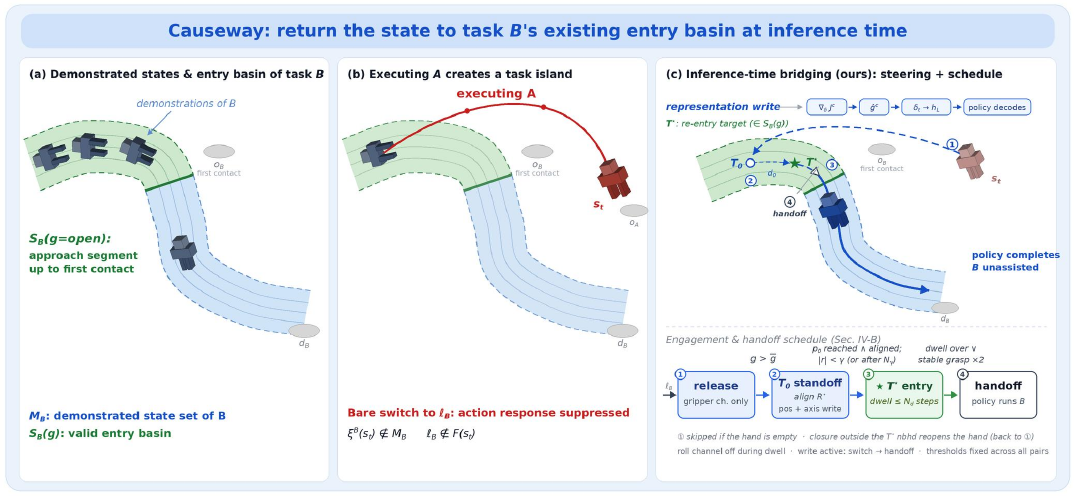}
\caption{\textbf{Causeway bridges task islands.}
(a)~Demonstrations of task $B$ define its demonstrated state set
$\mathcal{M}_B$, with $\mathcal{S}_B(g)$ denoting the valid
re-entry region.
(b)~Executing task $A$ can leave the robot in a state from which
task $B$ is no longer accessible, so a bare switch to $\ell_B$
fails.
(c)~Causeway steers the action-stream representation with a
state-directed gradient (top) until the robot re-enters the
basin, scheduled through four phases \circled{1}--\circled{4}
(bottom, Sec.~\ref{sec:flow}). Dashed: write active; solid: the
frozen policy completes $B$ on its own.}
\label{fig:causeway}
\end{figure*}


\section{Related Work}

\paragraph{VLA policies and instruction switching}
VLA policies use language to select behaviors from broad manipulation repertoires \cite{brohan2023rt2,kim2024openvla,black2024pi0,nvidia2025groot,octo2024,liufu2026repovlarecoverydrivenpolicyoptimization,tan2026rl2vlaadaptiverllatent}, while standard benchmarks generally evaluate one instruction from a prescribed initial-state distribution~\cite{liu2023libero}. High-level systems  compose skills using learned affordances~\cite{ahn2022saycan}, and interactive approaches use language corrections to redirect ongoing behavior~\cite{shi2024yay}. Recent works directly study task switching in VLA policies. SwitchVLA trains an execution-aware policy to select forward, rollback, and advance behaviors using phase and contact supervision~\cite{li2025switchvlaexecutionawaretaskswitching}. ReSteer formalizes state-dependent steerability, evaluates post-switch success across intermediate states, and improves switching through generated transition data and policy refinement~\cite{chen2026resteerquantifyingrefiningsteerability}. In contrast, Causeway shares the same observation that instruction responsiveness depends on state, but addresses a different recovery setting. Rather than training switching behaviors or fine-tuning the policy on generated transitions, Causeway uses target-relative geometry to restore access to a target task at inference time. This is consistent with the
initiation-set perspective in hierarchical reinforcement learning~\cite{sutton1999options,konidaris2009skill}. 


\paragraph{Inference-time steering and representation intervention}
Inference-time guidance modifies model predictions without updating parameters. Classifier-free and contrastive guidance amplify differences between conditional predictions~\cite{ho2022cfg,li2023contrastive}. In robotics, DeLock applies contrastive prompt guidance to the denoising dynamics of a post-trained VLA, using the difference between new and previously trained prompts to preserve responsiveness under novel instructions~\cite{delock}. Activation-steering methods instead add directions derived from contrasting examples or representations to intermediate model states~\cite{turner2023actadd,rimsky2024caa,zou2023repe}.
These approaches strengthen an instruction-conditioned signal but do not explicitly determine which physical state would make the target task executable. In contrast, Causeway derives its intervention from the error between the current end-effector pose and a demonstration-supported re-entry pose. 
Our objective is state recovery rather than amplification of the instruction contrast.


\paragraph{Recovery and low-level control}
Motion planning, Cartesian servoing\cite{chaumette2006servo}, and task-specific recovery policies can drive a robot directly toward a desired configuration. Such methods generate or override low-level actions outside the task policy. Recovery Chaining~\cite{vats2025recoverychaininglearninglocalrecovery} learns a recovery
policy that drives the system into a nominal controller's
initiation set~\cite{sutton1999options,konidaris2009skill};
Causeway performs the analogous re-entry by steering the frozen
policy's own representation, adding no policy. It uses the
re-entry pose only to define an action-space objective whose
gradient steers the representation---the frozen VLA still
generates every action---and thus lies between language-only
guidance and external control.


\section{Task Islands and Accessibility}
\label{sec:islands}

\subsection{Problem Formulation}

\label{sec:setup}
Let $\pi(a\mid x,\ell)$ be a language-conditioned policy over
observations $x$ and instructions $\ell$, trained by imitation on a set
of tasks $\{\ell_k\}$ with demonstration sets $\{\mathcal{D}_k\}$. Running
$\pi$ closed-loop under a fixed $\ell_k$ from the initial state
distribution induces a trajectory distribution $p_k^{\pi}(\tau)$; task $k$
succeeds when its goal predicate $G_k(\tau)$ holds at some step, and we
use the symbol $\rho_k = \Pr_{\pi}[G_k(\tau)=1]$ for the policy's success rate on
$k$ from the initial state.

In \emph{instruction switching}: the policy executes $\ell_A$ from
the initial state, and at some state $s$ along that execution the
instruction switches permanently to $\ell_B$. For such a state,
let
\begin{equation}
  \rho_{A\to B}(s) = \Pr_{\pi}\!\big[\,G_B(\tau)=1 \mid s,\ \ell_B\,\big]
  \label{eq:switch_rate}
\end{equation}
be the probability that the goal of $B$ is subsequently reached, with the from-scratch success $\rho_B$ as the reference. A \emph{task island} refers to a state $s$
at which $\rho_{A\to B}(s)$ is low, although $\rho_B$ is high: the policy is
competent at $B$, yet cannot begin it from $s$. Thus, a task island is defined relative to a target task, policy, and
rollout horizon. The remainder of this section makes this outcome-level definition
usable, and explains the failure it names.
Sec.~\ref{sec:state} introduces an object-centric lens---the state
$\xi$, the entry basin, and its measured tolerances---and uses it to
construct a computable geometric proxy for
Eq.~\eqref{eq:switch_rate}. Secs.~\ref{sec:escape}
and~\ref{sec:afterA} locate islands behaviorally---during execution
and after task completion---and tie their occurrence to the geometry
of $\xi$. Sec.~\ref{sec:gating} then asks \emph{why} a new instruction
alone does not escape an island, and \emph{what} remains steerable;
Sec.~\ref{sec:method} turns these answers into Causeway.


\begin{figure}[t]
  \centering
  \includegraphics[width=\linewidth]{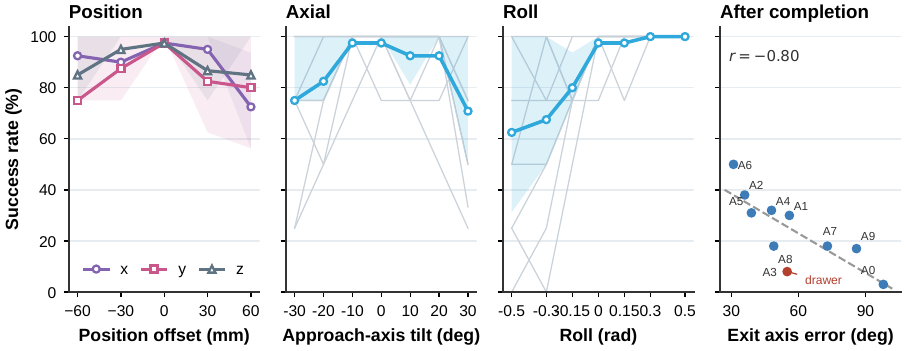}
  \caption{\textbf{Entry-basin tolerance and terminal-state mismatch.}
Left three panels: success under component-wise perturbations around
a re-entry pose; thin
lines show individual tasks and bands show interquartile ranges.
Right: bare-switch success after $A$ completes vs.\ the source's exit
approach-axis error ($r{=}-0.80$); A3 (red): low success despite moderate axis
error---its hand exits inside the drawer.}
  \label{fig:tolerance}
\end{figure}

\subsection{Object-Centric State and Entry Basin}
\label{sec:state}

\paragraph{Object-centric state}
Task $B$ manipulates an object $o_B$ and, for pick-and-place tasks,
moves it to a destination $d_B$. We represent the state relevant  to
$B$ as
\begin{equation}
  \xi^{B} = \big(p_{EE}-p_{r_B},\,R_{EE},\,g\big),\qquad
  r_B =
  \begin{cases}
    o_B & \text{before contact,}\\
    d_B & \text{after the grasp,}
  \end{cases}
  \label{eq:xi}
\end{equation}
where $p_{\mathrm{EE}}$ and $R_{\mathrm{EE}}$ are the world-frame end-effector position
and orientation; $p_{r_B}$ is the position of task reference $r_B$, and $g$
is the gripper state. Changing the reference from $o_B$ to $d_B$
keeps $\xi^B$ task-relative during approach and transport.
Orientation remains in the world frame
because object orientations are fixed in our evaluation settings. Consequently, the same
physical state maps to a different $\xi^B$ for each target task.

\paragraph{Demonstrated state set and entry basin}
Let $\mathcal{D}_B$ be the demonstrations of task $B$ and
$\mathcal{M}_B=\{\xi^B(s):s\in\mathcal{D}_B\}$ the
\emph{demonstrated state set} in $\xi$-space.
The \emph{entry basin} $\mathcal{S}_B(g)$ is the phase-compatible
subset selected by the gripper state $g$: the approach segment up
to first contact when the gripper is open, and the beginning of
the placement segment when it already holds $o_B$. We expand
these demonstrated states component-wise using the measured
position, approach-axis, and roll tolerances in
Fig.~\ref{fig:tolerance}; each component lies within tolerance of a demonstrated state in the
corresponding phase.
Thus, $\mathcal{S}_B(g)$ is a
demonstration-derived region of states compatible with entering
the corresponding phase of task $B$.
Direct placement at a demonstrated pre-grasp configuration
recovers task completion (Placement, Table~\ref{tab:main}). 

\paragraph{Executable set and task islands}
We operationalize the \emph{empirical executable set} of a state $s$ as the
instructions whose demonstrated state set contains it:
\begin{equation}
  F(s) = \{\, \ell : \xi^{\ell}(s) \in \mathcal{M}_\ell \,\}.
  \label{eq:executable}
\end{equation}
Because $g$ is a component of $\xi$, membership in
$\mathcal{M}_\ell$ already enforces gripper-phase compatibility.
For example, an open-gripper state cannot lie in a closed-gripper
segment, so Eq.~\eqref{eq:executable} requires no separate phase
conditions.
We treat a switch from $A$ to $B$ at state $s$ as viable when
$\ell_B\in F(s)$. Empirically, $\rho_{A\to B}(s)$ is low when
$\xi^B(s)\notin\mathcal{M}_B$, which gives our geometric
operationalization of a task island. As the policy executes $A$,
$\xi^\ell(s_t)$ leaves the demonstrated state sets of other tasks and
$|F(s_t)|$ generally contracts.

\begin{figure}[t]
\centering
\includegraphics[width=0.85\columnwidth]{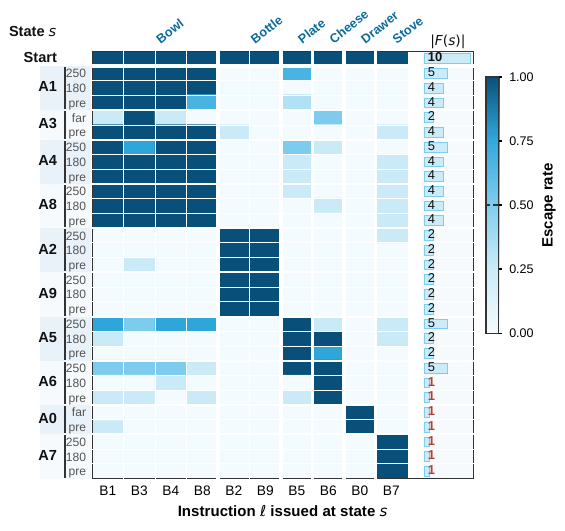}
\caption{\textbf{Escape map (OpenVLA-OFT, LIBERO-Goal).}
250, 180: hand--object distance,
mm; pre: pre-grasp; far/pre for the two drawer tasks. Columns: the instruction issued at that
state, grouped by target object. Cells: fraction of trials
reaching the object of $\ell$. Right: $|F(s)|$, the number of
instructions with escape rate $\geq0.5$; $|F(s)|{=}1$ (red) marks
committed states.}
\label{fig:escape}
\end{figure}

\paragraph{Validation of the entry basin}
The nominal rate $\rho_B$ from Sec.~\ref{sec:setup} serves as the
\emph{reference level}: any return is evaluated against this level,
not against 100\% success. When the end effector is placed directly at a pose in
$\mathcal{S}_B$ and $\ell_B$ is issued, the policy---which
conditions only on the current observation---recovers this level
while the scene remains near $B$'s demonstrated configuration;
completion of $A$ can alter the scene
(Table~\ref{tab:main}). The
same holds for poses sampled at different points along the
approach segment, showing that $\mathcal{S}_B$ has non-zero
extent and the return target is not a specific pose.

\paragraph{Tolerance bands of the basin}
Starting from a pose placed directly on the demonstrated approach (e.g.,  120\,mm) before contact, we perturb one component of $\xi$ at a
time and rerun the policy (Fig.~\ref{fig:tolerance}; 10 tasks
$\times$ 4 trials per point).
Success remains near the reference level for position offsets within
30\,mm, while the approach axis tolerates about $\pm20^\circ$. Roll
is asymmetric: $+0.5$\,rad shows no measurable loss, whereas
$-0.15$\,rad already costs 20\,pp. Gripper state is discrete: a
pre-grasp pose with a closed gripper lies outside $\mathcal{S}_B$.
These component-wise differences motivate use of separate write channels
and handoff criteria of Sec.~\ref{sec:method}.

\subsection{Islands under Mid-Execution Switching}
\label{sec:escape}
We next ask when $\ell_B$ remains executable as the policy performs $A$.

\begin{figure}[t]
\centering
\includegraphics[width=\columnwidth]{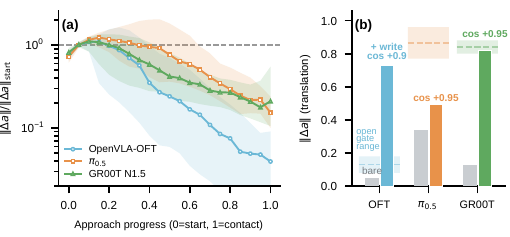}
\caption{\textbf{Instruction influence attenuates along the
approach, yet the gated representation remains steerable.}
(a)~Action response to swapping $\ell_A{\to}\ell_B$, relative to
the start of the approach (median, IQR). (b)~At post-approach
states, a gradient direction computed on the current forward pass
restores the response toward the open-gate range and aligns the
decoded motion with the return direction.}
\label{fig:gate}
\end{figure}

\paragraph{Measurement: the escape map}
We probe executability at a state $s$ by rolling the
policy out closed-loop for a fixed horizon under each instruction
$\ell$ and recording whether the end effector reaches its target
object. The resulting state-by-instruction table is the
\emph{escape map}; thresholding its escape rates provides a behavioral
estimate of the geometrically defined $F(s)$
(Fig.~\ref{fig:escape}). We sample states from the initial pose
through approach, pre-grasp, and grasp, and construct the detailed
map on OpenVLA-OFT because its instruction response attenuates most strongly along the approach (Fig.~\ref{fig:gate}a); cross-model
generality is evaluated in Sec.~V.
In all tested cases, short-horizon reachability agrees with full-length rollouts.

\paragraph{Object identity structures surviving tasks}
At the shared initial pose all ten instructions are executable;
as the policy executes $A$, most cross-object instructions fall
out of $F(s)$ (Fig.~\ref{fig:escape}).The remaining set is strongly structured
by object identity: at the final sampled state, all four bowl-source
tasks retain the four bowl instructions, while both bottle-source
tasks retain the two bottle instructions. Several singleton-object
sources reach $|F(s)|{=}1$, where only one instruction remains. The evolution need not be monotone:
along A3, $|F(s)|$ falls from 10 to 2 and then rises to 4 as the
remaining bowl instructions cross the escape threshold again. This
rebound shows that executability follows the state rather than
elapsed time. The same-object case is followed beyond
pre-grasp below.

\paragraph{Same-object pairs}
\label{sec:sameobj}
When $A$ and $B$ manipulate the same object ($o_A{=}o_B$), post-grasp viability depends on whether their demonstrated grasp and early-transport states overlap. A switch can survive the grasp while those states remain compatible, then fail as the transport paths diverge. Without such overlap, the held state may become an island immediately.

\subsection{Islands after Task Completion}
\label{sec:afterA}

The terminal state induced by $A$ can itself be an island for $B$.
Across source tasks, bare-switch success after completion ranges from
3\%
to 50\% and is strongly negatively associated with the mean
target-relative approach-axis error at exit, which spans
$31^\circ$--$98^\circ$ ($r{=}-0.80$;
Fig.~\ref{fig:tolerance}). A3 is an outlier because the hand ends inside the drawer region despite a mid-range axis error; A7
also ends with the gripper closed. Terminal mismatch can therefore involve several components of $\xi^B$, motivating the orientation channels and release step in
Sec.~\ref{sec:method}.

\subsection{The Representation Gate}
\label{sec:gating}

The preceding analysis locates task islands. We now examine why a new instruction alone may not escape them.

\paragraph{State attenuates the instruction's effect on action}
At a fixed observation along $A$'s trajectory, we run the policy
with $\ell_A$ and $\ell_B$ and measure the action response
$\|\Delta a\|$ as the norm of the mean translational difference
between the two decoded action chunks. The new instruction always changes the text-token representation, but its effect on action depends on state.
Before contact, $\|\Delta a\|$ drops by one to two orders of
magnitude on all three architectures
(Fig.~\ref{fig:gate}a)---beyond what converging near-optimal
actions explain, since a gradient direction at the same states
still produces motion toward $B$ (Fig.~\ref{fig:gate}b); we call this state-conditioned attenuation the \emph{gate}.
It appears along the same approach over which F(s) changes (Sec.~\ref{sec:escape}), linking the behavioral and representational findings.

\paragraph{The action representation remains steerable at gated states}
When the instruction no longer changes the action, a gradient from the current forward pass restores the action response and aligns the decoded motion with the return direction on all three architectures (Fig.~\ref{fig:gate}b). This result identifies both the recovery target and intervention point: steer the action stream until $\xi^B(s)$ returns to
$\mathcal{S}_B(g)$ (Sec.~\ref{sec:method}).

\section{Bridging Task Islands by Representation Steering}
\label{sec:method}

Section~III suggests three design choices: return to $\mathcal{S}_B(g)$, intervene in the action-stream representation, and use a state-directed gradient when the instruction's influence is attenuated. Causeway realizes all three: at test time it consumes only the
robot's proprioceptive state and a re-entry pose $T^\star$
precomputed from the demonstrations of $B$, and every action is
decoded by the frozen policy. A \emph{representation write} moves the decoded action toward a target pose in $\mathcal{S}_B(g)$ (Sec.~\ref{sec:write}), while a proprioceptive rule controls \emph{engagement and handoff} (Sec.~\ref{sec:flow}). Separate channels handle position, approach axis, roll, and gripper state because each has a different re-entry tolerance.




\subsection{Representation Write}
\label{sec:write}


\paragraph{Write location}
For each architecture, we write within the action stream identified
in Sec.~\ref{sec:gating}: one layer $L$ on $\pi_{0.5}$ and
GR00T N1.5, and multiple layers $\mathcal{L}$ on OpenVLA-OFT.
Below, $L$ denotes a generic write location; on OpenVLA-OFT the same
operation is applied at each $L\in\mathcal{L}$. Let
$h_L^{(k)}\in\mathbb{R}^{N\times D}$ denote the hidden state of the
$N$ action tokens at layer $L$ during decoding step $k$, where $D$
is the hidden dimension and $k=1,\ldots,K$. The final decoded action
chunk is
\begin{equation}
    a = f_L\!\left(h_L^{(1)},\ldots,h_L^{(K)}\right),
    \qquad a\in\mathbb{R}^{H\times 7},
    \label{eq:decode}
\end{equation}
where $f_L$ denotes the remaining decoding computation from $L$ to
the final action, $H$ is the action-chunk horizon, and each decoded action comprises translation and rotation increments
($\mathbb{R}^3$ each) and one gripper command.
For iterative action heads, $K>1$; for a parallel-decoding head,
$K=1$ and Eq.~\eqref{eq:decode} reduces to a single forward pass.

\paragraph{Target}
$T^\star=(p^\star,R^\star)$ is chosen from the phase-compatible
segment of $\mathcal{S}_B(g)$ by the same fixed rule for all task
pairs and models, without pair-specific tuning. A local neighborhood of $T^\star$ defines the handoff condition
(Sec.~\ref{sec:flow}).

\nocite{tan2026rl2vlaadaptiverllatent,liufu2026repovlarecoverydrivenpolicyoptimization}

\paragraph{Return geometry}
Let $(p,R)$ denote the current end-effector pose, corresponding to
$(p_{EE},R_{EE})$ in Eq.~\eqref{eq:xi}. The return geometry toward
$T^\star$ is computed by the translational and rotational offsets
\begin{equation}
    e^{p}=p^\star-p,
    \qquad
    e^{r}=\log(R^\star R^{-1})^{\vee},
    \label{eq:poserr}
\end{equation}
where $(\cdot)^\vee$ maps a skew-symmetric matrix in
$\mathfrak{so}(3)$ to its corresponding three-dimensional rotation
vector. Both errors are expressed in the world frame. We compute the
rotational error from the relative rotation $R^\star R^{-1}$,
consistent with the policy's left-multiplicative rotation increment.

\paragraph{Gradient write}
Let $\delta\in\mathbb{R}^{D}$ denote a shared representation
perturbation. At each decoding step $k$, the same perturbation is
broadcast to all $N$ action tokens at layer $L$,
\begin{equation}
    \tilde h_L^{(k)} = h_L^{(k)} + \mathbf{1}_N\delta^\top,
    \qquad k=1,\ldots,K,
    \label{eq:perturbed_hidden}
\end{equation}
where $\mathbf{1}_N\in\mathbb{R}^{N}$ is the all-ones vector. The
same $\delta$ is shared across all decoding steps and, on
OpenVLA-OFT, across every write layer $L\in\mathcal{L}$. Let
$a(\delta)$ denote the final action chunk produced by the complete
decoding process under all such injections.
Each channel $c$ defines a scalar objective $J^{c}(a)$ as the
component of the predicted motion along its desired direction
(e.g., $u^p$ for position, defined below). The corresponding
representation-space direction is
\begin{equation}
    \hat g^{c} = \left.
    \frac{\nabla_{\delta}J^{c}(a(\delta))}
         {\lVert\nabla_{\delta}J^{c}(a(\delta))\rVert}
    \right|_{\delta=0}.
    \label{eq:channel_grad}
\end{equation}
Each channel is normalized separately, so relative strengths are
set only by the per-channel gains introduced below. The gradient
is computed through the complete decoding computation. On
OpenVLA-OFT's parallel-decoding head, this reduces to a single
forward--backward pass.

\subsubsection{Channel Decomposition}
\label{sec:channels}

Let the $h$-th decoded action be
$a_h=(a_{h,xyz},a_{h,rot},a_{h,g})$, for $h=1,\ldots,H$.
We aggregate each component over the action chunk as
\begin{equation}
    \bar a_{xyz}=\sum_{h=1}^{H}a_{h,xyz},\qquad
    \bar a_{rot}=\sum_{h=1}^{H}a_{h,rot},\qquad
    \bar a_g=\sum_{h=1}^{H}a_{h,g}.
    \label{eq:chunk_sum}
\end{equation}
Let $\hat z_w=R\hat z$ denote the current gripper approach axis
in the world frame, where $\hat z$ is the unit approach axis in
the gripper frame. We define separate objectives for translation,
approach-axis alignment, roll, and gripper motion.

\paragraph{Position channel}
Let $u^p=e^p/\lVert e^p\rVert$ denote the unit direction toward
$p^\star$. We define
\begin{equation}
    J^p=\langle \bar a_{xyz},u^p\rangle.
    \label{eq:Jp}
\end{equation}
Substituting $J^p$ into Eq.~\eqref{eq:channel_grad} gives
$\hat g^p$, the representation direction that locally increases
decoded motion toward $p^\star$. The remaining channels are
constructed in the same way from their corresponding objectives.

\paragraph{Orientation decomposition}
We decompose the rotational error $e^r$ into a component that
changes the gripper approach direction and a component about that
direction:
\begin{align}
    e^\perp &= e^r-(e^r\!\cdot\!\hat z_w)\hat z_w,
    \label{eq:eperp}\\
    r &= e^r\!\cdot\!\hat z_w .
    \label{eq:roll}
\end{align}
Here, $e^\perp$ changes the approach direction and $r$ is the
signed roll error about the current approach axis.

\paragraph{Approach-axis channel}
Let $u^\perp=e^\perp/\lVert e^\perp\rVert$. We define
\begin{equation}
    J^\perp=\langle\bar a_{rot},u^\perp\rangle,
    \label{eq:Jperp}
\end{equation}
which yields the representation direction $\hat g^\perp$.

\paragraph{Roll channel}
The desired roll direction is
$\operatorname{sign}(r)\hat z_w$, giving
\begin{equation}
    J^r=
    \langle\bar a_{rot},
    \operatorname{sign}(r)\hat z_w\rangle.
    \label{eq:Jroll}
\end{equation}
The resulting representation direction is $\hat g^r$.
Because successful re-entry is sensitive to roll error
(Sec.~\ref{sec:islands}), we taper its gain near zero:
\begin{equation}
    \alpha_r(r)=\alpha_r^{\max}
    \min\!\left(1,\frac{|r|}{\kappa}\right)
    \mathbb{1}[|r|>\epsilon],
    \label{eq:taper}
\end{equation}
where $\kappa$ defines the proportional band and $\epsilon$ the
dead-band threshold.

\paragraph{Gripper channel}
When the hand holds $o_A$ and $o_A\neq o_B$, the return begins by
opening the gripper: $J^g=\bar a_g$ on the gripper command
channel, which yields $\hat g^g$. If the hand already holds
$o_B$, the closed-gripper branch of $\mathcal{S}_B(g)$ is used
instead.

\paragraph{Composition}
The write applied at inference step $t$ is
\begin{equation}
\delta_t=
\begin{cases}
    \alpha_g\hat g^g,
        & \text{during release},\\
    \alpha_p\hat g^p+
    \alpha_\perp\hat g^\perp+
    \alpha_r(r)\hat g^r,
        & \text{otherwise}.
\end{cases}
\label{eq:write_delta}
\end{equation}
At each inference step, the channel directions are evaluated at
$\delta=0$, combined according to Eq.~\eqref{eq:write_delta}. The final action is decoded by the policy from the shifted
representation (gain values in Sec.~\ref{sec:setup_exp}).

\subsection{Engagement and Handoff}
\label{sec:flow}

The representation write is scheduled from proprioception and the target pose
$T^\star=(p^\star,R^\star)$. Let $\hat z_w^\star=R^\star\hat z$ and we define
\begin{equation}
    T_0=(p_0,R^\star),\qquad
    p_0=p^\star-d_0\hat z_w^\star .
    \label{eq:prepoint}
\end{equation}

\paragraph{Engagement and return}
The write starts at the instruction switch. If the hand holds
$o_A\neq o_B$, only the gripper channel is active until the hand
opens beyond $\bar g$; otherwise, the position and orientation
channels engage immediately. They first drive the end effector
toward $T_0$ while aligning with $R^\star$. Once $p_0$ is reached
and the approach axis is aligned, the position target switches to
$p^\star$ when $|r|<\gamma$, or after $N_\gamma$ additional
steps. Upon entering the neighborhood of $T^\star$, the write
continues for at most $N_d$ steps; the roll
channel is disabled during this dwell to avoid constraining the
policy's subsequent wrist motion.

\paragraph{Handoff}
The write disengages after the dwell or earlier when the gripper
remains closed around a non-empty grasp for two consecutive steps
inside the neighborhood of $T^\star$. A closure outside this
neighborhood does not trigger handoff; the gripper channel reopens
the hand and the return continues.

This schedule determines only \emph{when} the write is active and
\emph{which} channels engage; it issues no motor commands; every
action, during re-entry and after handoff, is decoded by the
frozen policy.

\section{Simulation Experiments}
\label{sec:exp}

We test four hypotheses: task islands occur across models and switch timings (H1); Causeway restores access through re-entry (H2); neither instruction contrast nor action-level steering is sufficient (H3); the state-compatibility account extends beyond the main cross-object setting (H4).


\subsection{Setup}
 \label{sec:setup_exp}

\paragraph{Benchmarks}
We evaluate LIBERO-Goal's 90 ordered task pairs: the main evaluation
uses the 71 cross-object pairs, and the 14 same-object pairs serve
H4. Five pairs are excluded because executing $A$ displaces $B$'s
target configuration. We also evaluate LIBERO-Object to test
generalization beyond LIBERO-Goal.

\paragraph{Policies}We evaluate benchmark-fine-tuned checkpoints of OpenVLA-OFT~\cite{kim2025oft} (parallel decoding), $\pi_{0.5}$~\cite{pi05_2025} (flow matching), and GR00T~N1.5~\cite{nvidia2025groot} (diffusion). Writes target the action-stream layers diagnosed in Sec.~\ref{sec:gating}: layers $\{8, 12, 16, 20, 24, 28, 30\}$ for OpenVLA-OFT, layer 15 for $\pi_{0.5}$, and DiT block 16 for GR00T. OpenVLA-OFT and GR00T use a single gain profile across all pairs. For $\pi_{0.5}$, three fixed geometry-driven profiles are used: grasp-and-place targets use a larger translation gain, while sources A0 and A9 use a stronger orientation gain to counter post-completion wrist-angle errors (Sec.~\ref{sec:afterA}). All gain profiles remain fixed across pairs and switch timings, and
the schedule thresholds ($\gamma=0.3$~rad, $N_\gamma=20$,
$N_d=15$ steps; Sec.~\ref{sec:flow}) are additionally shared across
all three models.

\paragraph{Protocol and metric}Policies execute $\ell_A$ from benchmark initial states before switching to $\ell_B$ at step 30 (approach), step 60 (object held in 58--61\% of trials), or post-completion. Evaluating 71 pairs across 3 switch conditions, 8 initial states, and 3 rollouts yields $71\times3\times8\times3 = 5{,}112$ trials per model. Because switch states vary, success is evaluated over the state distribution $p_A^\pi$ induced by task $A$ at switch timing $\tau \in \{30, 60, \mathrm{done}\}$:$$\rho^{(\tau)}_{A\to B} = \mathbb{E}_{s_\tau\sim p_A^\pi} [\rho_{A\to B}(s_\tau)],$$under standard LIBERO criteria. All models achieve near-perfect from-scratch success on target tasks from clean initial states. 

\paragraph{Inference cost}
On one A100-80GB, per-chunk latency rises $2.5$--$2.8\times$
(102/75/128 $\to$ 253/187/362~ms; replanning 4.0/5.3/2.8~Hz;
$+{\leq}4.3$~GiB peak) on OFT/$\pi_{0.5}$/GR00T, active only
during re-entry.

\paragraph{Baselines}We compare against four steering baselines: (1) \emph{Bare switch}, which directly replaces $\ell_A$ with $\ell_B$; (2) \emph{Contrastive prompt guidance (CPG)}~\cite{delock}, which combines predictions via $v_A + w(v_B - v_A)$; (3) \emph{Contrastive activation addition (CAA)}~\cite{rimsky2024caa}, which injects online differences $d_L \propto \bar h_L(x_t, \ell_B) - \bar h_L(x_t, \ell_A)$ at each step through Causeway's write machinery; and (4) \emph{Action-space steering}, which applies the state-directed signal
and schedule as a Cartesian residual directly on the decoded
actions, evaluated on $\pi_{0.5}$, where the baseline signal is
strongest and floor effects are smallest.

\paragraph{References with Privileged recovery control}
We also report references with privileged recovery control on OpenVLA-OFT.
\emph{State placement} sets the robot to a demonstration pre-grasp
configuration of $B$ at the switch while preserving the scene. The
\emph{single-point} and \emph{wrist-first} servos drive the arm to
the same configuration with an external controller, the latter
along a hand-designed wrist-first path. These are references rather
than baselines: they rely on state manipulation, external control,
or hand-designed motion that Causeway does not use.

\begin{table}[t]
\centering
\caption{\textbf{Switch success (\%).} $\ell_B$ issued at Step 30,
Step 60, or after $A$ completes. \emph{Reach}: fraction reaching the
handoff neighborhood of $T^\star$; \emph{Fin$\mid$R.}: completion
given reach. Gray: references using external state manipulation. Best steering method per block in bold.}
\label{tab:main}
\setlength{\tabcolsep}{3pt}\renewcommand{\arraystretch}{0.95}
\small
\resizebox{\columnwidth}{!}{%
\begin{tabular}{@{}llcccccc@{}}
\toprule
 & & \multicolumn{4}{c}{Switch point} &
 \multicolumn{2}{c}{Re-entry to $T^\star$} \\
\cmidrule(lr){3-6}\cmidrule(l){7-8}
Model & Method & Step 30 & Step 60 & After & Avg.
 & Reach & Fin$\mid$R. \\
\midrule
\multirow{5}{*}{$\pi_{0.5}$}
 & Bare & 29.9 & 22.9 & 25.1 & 26.0 & 18.7 & 58.7 \\
 & CPG  & 35.2 & 21.1 & 26.5 & 27.6 & --   & --   \\
 & CAA  & 26.9 & 21.7 & 26.4 & 25.0 & -- & -- \\
 & Action & 37.5 & 21.3 & 26.6 & 28.5 & -- & -- \\
 & \cellcolor{oursrow}\textbf{Ours}
   & \cellcolor{oursrow}\textbf{60.9} & \cellcolor{oursrow}\textbf{60.4}
   & \cellcolor{oursrow}\textbf{63.4} & \cellcolor{oursrow}\textbf{61.6}
   & \cellcolor{oursrow}\textbf{61.0} & \cellcolor{oursrow}\textbf{74.0} \\
\midrule
\multirow{4}{*}{\shortstack[l]{OpenVLA-\\OFT}}
 & Bare & 5.1  & 2.9  & 2.2  & 3.4  & 9.3  & 25.3 \\
 & CPG  & 10.3 & 6.7  & 4.0  & 7.0  & --   & --   \\
 & CAA & 0.0 & 0.0 & 2.5 & 0.8 & -- & -- \\
 & \cellcolor{oursrow}\textbf{Ours}
   & \cellcolor{oursrow}\textbf{77.2} & \cellcolor{oursrow}\textbf{60.5}
   & \cellcolor{oursrow}\textbf{55.9} & \cellcolor{oursrow}\textbf{64.5}
   & \cellcolor{oursrow}\textbf{81.3} & \cellcolor{oursrow}\textbf{73.7} \\
\midrule
\multirow{4}{*}{GR00T N1.5}
 & Bare & 14.6 & 11.8 & 3.4  & 9.9  & 8.7  & 45.0 \\
 & CPG  & 13.2 & 10.0 & 7.7  & 10.3 & --   & --   \\
 & CAA & 5.0 & 5.0 & 3.8 & 4.6 & -- & -- \\
 & \cellcolor{oursrow}\textbf{Ours}
   & \cellcolor{oursrow}\textbf{67.2} & \cellcolor{oursrow}\textbf{43.3}
   & \cellcolor{oursrow}\textbf{32.0} & \cellcolor{oursrow}\textbf{47.5}
   & \cellcolor{oursrow}\textbf{79.0} & \cellcolor{oursrow}\textbf{59.3} \\
\midrule
\multirow{3}{*}{\shortstack[l]{\emph{References using }\\\emph{external state manipulation}}}
 & \cellcolor{privrow}\emph{Servo (single-point)}
   & \cellcolor{privrow}67.5 & \cellcolor{privrow}49.2
   & \cellcolor{privrow}25.0 & \cellcolor{privrow}47.2
   & \cellcolor{privrow}--   & \cellcolor{privrow}-- \\
 & \cellcolor{privrow}\emph{Servo (wrist-first)}
   & \cellcolor{privrow}97.0 & \cellcolor{privrow}76.0
   & \cellcolor{privrow}48.0 & \cellcolor{privrow}73.7
   & \cellcolor{privrow}--   & \cellcolor{privrow}-- \\
 & \cellcolor{privrow}\emph{Placement}
   & \cellcolor{privrow}95.0 & \cellcolor{privrow}93.1
   & \cellcolor{privrow}49.4 & \cellcolor{privrow}79.2
   & \cellcolor{privrow}--   & \cellcolor{privrow}-- \\
\bottomrule
\end{tabular}}
\end{table}

\subsection{Results}

\begin{figure}[t]
  \centering
  \includegraphics[width=\columnwidth]{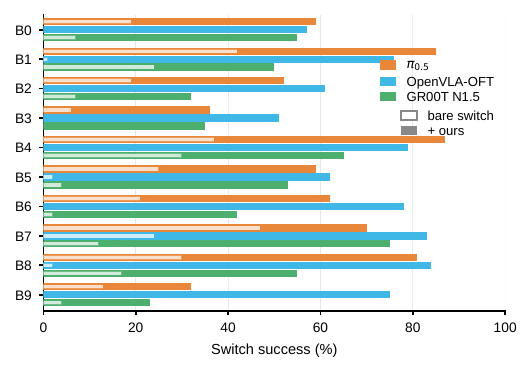}
  \caption{Per-target switch success, pooled over source tasks,
switch timings, initial states, and rollouts. Filled: with the
write; white inset: bare switch. Bottom: pooled average.}
  \label{fig:taskdim}
\end{figure}

\begin{table}[t]
\centering
\caption{\textbf{Real-robot instruction switching} (xArm, $\pi_{0.5}$ fine-tuned). Success rate (\%) over 5 trials per cell. $\ell_B$ is
issued during the approach, at the pre-grasp pose, or after $A$
completes. Tasks: press \emph{button}, put the bowl on the \emph{plate} /
\emph{shelf}, put the corn in the drawer (\emph{corn}). $\dagger$: same-object pairs (both move the bowl).
\emph{Solo}: the policy's own success on $B$ from the standard
start.}
\label{tab:realrobot}
\setlength{\tabcolsep}{2.5pt}
\renewcommand{\arraystretch}{0.95}
\footnotesize
\begin{tabular*}{\columnwidth}{@{\extracolsep{\fill}}l cc cc cc@{}}
\toprule
 & \multicolumn{2}{c}{Approach} & \multicolumn{2}{c}{Pre-grasp}
 & \multicolumn{2}{c}{After $A$} \\
\cmidrule(lr){2-3}\cmidrule(lr){4-5}\cmidrule(l){6-7}
$A \to B$ & bare & \textbf{ours} & bare & \textbf{ours} & bare & \textbf{ours} \\
\midrule
button $\to$ shelf   & \textbf{100} & \textbf{100} & 0   & \textbf{100} & 40 & \textbf{80}  \\
shelf $\to$ button   & 0   & \textbf{100} & 0   & \textbf{80}  & 0  & \textbf{60}  \\
button $\to$ plate   & 0   & \textbf{100} & 0   & \textbf{100} & 0  & \textbf{100} \\
plate $\to$ button   & 0   & \textbf{100} & 0   & \textbf{80}  & 0  & \textbf{100} \\
corn $\to$ plate     & 0   & \textbf{100} & 0   & \textbf{100} & 80 & \textbf{100} \\
\addlinespace[2pt]
plate $\to$ shelf$^\dagger$ & \textbf{100} & \textbf{100} & 0   & \textbf{100} & 0 & \textbf{60} \\
shelf $\to$ plate$^\dagger$ & \textbf{100} & \textbf{100} & \textbf{100} & \textbf{100} & 0 & \textbf{80} \\
\midrule
Pooled & 42.9 & \textbf{100} & 14.3 & \textbf{94.3} & 17.1 & \textbf{82.9} \\
\midrule
\multicolumn{7}{@{}l}{\emph{Solo} (\%): button 80, plate 100, shelf 100, corn 80.}\\
\bottomrule
\end{tabular*}
\end{table}

\paragraph{H1: Causeway bridges task islands}
Across 71 cross-object pairs and three switch timings, bare-switch success
averages 26.0\% for $\pi_{0.5}$, 3.4\% for OpenVLA-OFT, and 9.9\% for
GR00T (Table~I), despite near-perfect from-scratch performance. Causeway
raises success to 61.6\% ($+35.6$ percentage points), 64.5\% ($+61.1$),
and 47.5\% ($+37.6$), respectively. Gains are positive for every model,
timing, and target (Fig.~\ref{fig:taskdim}), ranging from $+28.6$ points for GR00T after
completion to $+72.1$ for OpenVLA-OFT at step 30.

\begin{figure*}[!t]
  \centering
  \includegraphics[width=0.8\textwidth]{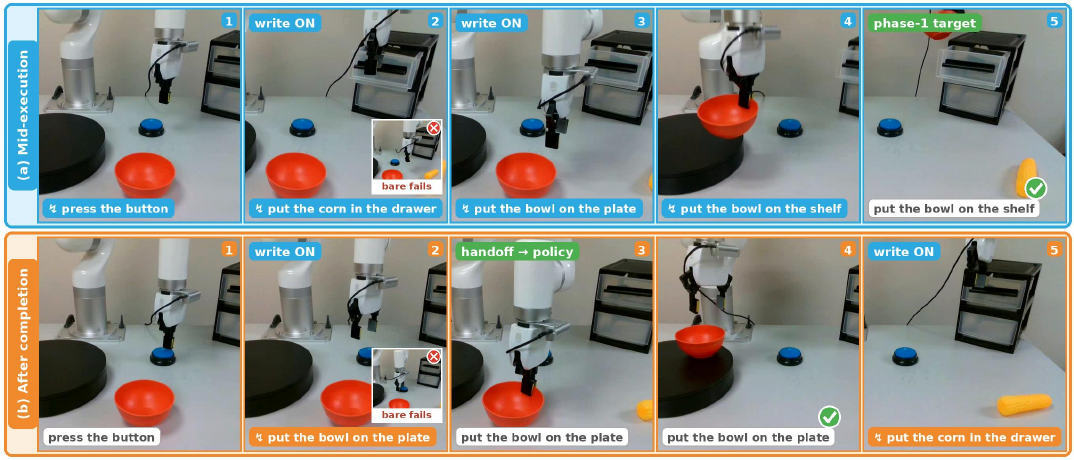}
  \caption{\textbf{Real-robot chained switching} (xArm, $\pi_{0.5}$, fine-tuned then frozen): \emph{Top Row:} Each instruction interrupts the previous task
  mid-execution; the final switch re-routes the held bowl from the
  plate to the shelf via the phase-conditioned target
  (closed-gripper branch, Sec.~\ref{sec:flow}); \emph{Bottom Row:} Instructions
  issued after each task completes. Solid instruction chips mark
  the frame where a new instruction takes effect; insets show the
  bare switch failing from the same state. Quantitative pairwise results in
  Table~\ref{tab:realrobot}.}
  \label{fig:realrobot}
\end{figure*}

\paragraph{H2: Causeway restores access through re-entry}
Causeway raises the rate of reaching the handoff neighborhood by 42--72 percentage points
 and increases completion conditioned on reaching it for
all three models (Table~\ref{tab:main}, Fin$\mid$R). Thus, it both reaches the neighborhood more
often and arrives in states from which the frozen policy completes $B$
more reliably. $T^\star$ is a representative target within the entry
basin, not a uniquely tuned recovery pose (Section~III-B).

\paragraph{H3: neither instruction contrast nor action-level
steering overcomes the gate}
Prompt-level contrast provides no consistent recovery: CPG shifts
average success by at most $+3.6$~pp and reduces it at some
timings (Table~\ref{tab:main}). Representation-level contrast
fails more decisively. CAA uses the same write machinery as
Causeway and differs only in direction, yet performs at or below the bare switch on all three models on
average (Table~\ref{tab:main}). Consistently, at gated states the gradient write induces
target-aligned motion while the instruction-difference direction
does not (Fig.~\ref{fig:gate}b).

The same state-directed signal also performs poorly when applied
only at the action level. On $\pi_{0.5}$, a Cartesian residual
improves the step-30 switch but falls near bare
at step~60 and after completion, while the representation write exceeds 60\%
at all three timings. Each alternative therefore lacks at least
one of two ingredients: state-directed steering or
representation-level injection.

\paragraph{Privileged recovery references}
On OpenVLA-OFT, Causeway (64.5\%) outperforms even the
externally controlled single-point servo (47.2\%). Stronger
privileged references perform better: the hand-designed
wrist-first servo reaches 73.7\%, and direct state placement
79.2\% (Table~\ref{tab:main}). Thus, matching or exceeding
Causeway requires additional external control, hand-designed
motion, or direct state manipulation.

\paragraph{H4: state compatibility beyond the main setting}
\label{sec:h4}
Same-object switches follow the state-compatibility account of
Sec.~\ref{sec:escape}: switches remain viable while the two
tasks share compatible holding states and fail after their
trajectories diverge. On LIBERO-Object, the write lifts the
committed-state switch from 0\% to 57.5\% on the evaluated
pairs, extending the recovery beyond the main cross-object
setting.

\section{Real-World Experiments}
\label{sec:real}

We validate Causeway on an xArm~6 with $\pi_{0.5}$, fine-tuned on
30 teleoperated demonstrations per task and frozen at inference;
re-entry poses $T^\star$ are extracted by
the rule of Sec.~\ref{sec:write}. We evaluate four tasks—\textbf{button}, \textbf{plate}, \textbf{shelf}, and \textbf{corn} (80–100\% solo success; Table~\ref{tab:realrobot})—across 7 ordered pairs (5 cross-object, 2 same-object). Target instructions $\ell_B$ are issued at three transition points: mid-approach, at pre-grasp, or post-completion (5 trials/cell).
\paragraph{Islands persist and Causeway bridges them}Bare-switch
success collapses as the robot approaches the object---from 42.9\%
mid-approach to 14.3\% and 17.1\% at pre-grasp and post-completion
(Table~\ref{tab:realrobot})---due to policy stalling or continuing
toward the source object (Fig.~\ref{fig:realrobot}). Causeway
elevates pooled success to 82--100\%; every
cell reaches at least 60\%, and zero-success pairs recover to
80--100\%.
\paragraph{Held objects and chained switching}For same-object pairs with an already-held item, Causeway uses the closed-gripper branch of $\mathcal{S}_B(g)$ (Sec.~\ref{sec:flow}) to route directly to the target placement segment—e.g., re-routing a held bowl from plate to shelf without put-down or re-grasp (Fig.~\ref{fig:realrobot}a). Across chained task sequences, Causeway enables continuous language-driven redirection mid-execution or post-completion without resets, whereas bare switches fail.

\section{Conclusion, Limitations, and Future Work}
We characterize task islands---states from which a competent policy
cannot begin a new task---and introduce Causeway, a training-free, inference-time method that applies a representation write by back-propagating through a frozen policy's decoding during return motion. Across three VLA architectures on LIBERO-Goal, Causeway elevates bare-switch success from 3–26\% to 47–65\%, and
transfers to LIBERO-Object and a physical xArm. Causeway depends on task-relative coordinates ($\xi$) and demonstration coverage, with performance degrading under unmodeled factors (such as large scene shifts or in-grasp dynamics) and requiring per-architecture calibration. Future work will use these diagnostics to embed capability accessibility directly into policy training.



\bibliographystyle{IEEEtran}
\bibliography{refs}
\end{document}